# Dissecting Geometric Constraints in Grokking: Interaction-Dependent Acceleration in Modular Arithmetic

**Zachary A. Hampton[1], Wasiq Khan[2] (SMIEEE, Ph. D), and Bilal Khan[1], Ph.D.**

[1]School of Computer Science and Engineering, California State University, San Bernardino, {Zachary.Hampton, Bilal.Khan}@csusb.edu

[2]School of Computer Science and Mathematics, Liverpool John Moores University, UK, w.khan@ljmu.ac.uk

## Abstract

Spherical transformer configurations can accelerate grokking through simultaneous modifications of several model properties. The present study investigated the contributions of these modifications in the published Sphere+Fourier configuration using 960 controlled modular-arithmetic runs. A complete factorial analysis varied residual projection, readout weight normalization, logit scale, and weight decay. Across four matched settings with weight decay, the successful-run mean crossing times without projection were 8-37 times those observed with projection. However, residual projection was insufficient to induce grokking without additional interventions. Various other unprojected configurations also generalized. Since projection altered the magnitude entering the readout, these comparisons could not separate geometry from output scale. A separate 100-run fixed-horizon study demonstrated faster first crossing for the complete configuration across five train/test partitions. Faster crossing did not necessarily correspond to greater persistence over extended training. Control experiments demonstrated rapid grokking without Fourier initialization. Task and depth sweeps also identified conditions in which the configuration failed to reach the criterion by 50K epochs. Accordingly, the observed acceleration depended on interactions among the components and did not extend uniformly across the tested tasks or depths.

## 1. Introduction and Related Work

Grokking describes delayed generalization in which a neural network first fits its training set without improving held-out performance beyond near-chance accuracy. Longer training can subsequently produce generalization (Power et al., 2022). Modular arithmetic provides a useful, controlled setting for investigating this delay since its finite operation table permits exact evaluation of every held-out input. Previous studies have examined Fourier-like circuits, competition between memorizing and generalizing subnetworks, kernel-to-feature transitions, and the influence of regularization and representation learning (Nanda et al., 2023; Merrill et al., 2023; Varma et al., 2023; Mohamadi et al., 2024; Liu et al., 2022a, 2022b).

Architectural and optimization choices can reduce the delay preceding generalization. Yıldırım (2026) introduced spherical residual streams with cosine unembedding and demonstrated accelerated grokking on cyclic modular tasks. The study also examined Fourier initialization, conventional normalization, learning rates, weight decay, uniform attention, multiplication, task alignment, and multiple seeds. Soft radial suppression accelerated generalization under a different target radius and intervention site (Tiwari et al., 2026). Normalization placement and optimization choices have also been shown to influence grokking (Singh et al., 2026). The present study extended these investigations through analysis of component interactions and their limitations.

The specific question addressed in the present study concerned the components responsible for acceleration in a spherical configuration. Sphere+Fourier combines residual projection, a normalized readout, a larger logit scale, Fourier initialization, and removal of weight decay. Comparison with the baseline therefore changes these properties simultaneously. Resolving the observed acceleration required systematic evaluation of their combinations to identify the conditions under which projection was beneficial or ineffective.

Accordingly, a complete $2^4$ factorial experiment examined component interactions with Fourier initialization disabled. Separate initialization controls connected the no-Fourier projected configuration to the complete Sphere+Fourier configuration. The complete configuration for new train/test splits was then evaluated with independent-partition study, with the boundaries examined via the task and depth sweeps. Therefore, the component interactions, robustness, and limits of accelerated grokking were comprehensively addressed via the 960-run experiment.

# 2. Methods

## 2.1. Study design and outcomes

The factorial experiment was conducted to resolve interactions among individual components. A separate 100-run study compared the complete baseline and Sphere+Fourier configurations across data partitions, model seeds, and shallow depths. Additional controls examined normalization, logit scale, Fourier features, tasks, and depth. Token embeddings were compared at matched training ages in a separate 20-run study. Complete protocols, detailed tables, and an inventory of all 13 experimental groups and 960 runs are provided in the Supporting Information (**Sections S1-S5; Supplementary Material**). Held-out classification accuracy was evaluated every 200 epochs and at epoch 1, except in the soft-penalty sweep. Fixed-horizon experiments always included epoch 50,000. The grok epoch was defined as the first observed checkpoint with test accuracy strictly above 0.95. This outcome was preregistered for the independent-partition study. Checkpoint-based evaluation left the exact transition time unknown. Runs without an observed crossing were administratively censored at 50K. Outcomes included crossing counts, the conditional mean ± sample standard deviation among successful runs, and restricted mean crossing time through 50K (RMST-50K). RMST was calculated across all runs as the mean of min(T,50,000), assigning 50,000 to runs without a crossing. Differences among the fastest configurations were interpreted within the resolution of the 200-epoch evaluation grid (Section S1.3, Supporting Information). The post hoc trajectory analysis preserved the preregistered first-crossing outcome. A recorded relapse was defined as any subsequent checkpoint at or below 0.95 following the first crossing. Horizon persistence required final test accuracy above 0.95. Representation measurements at matched training ages were interpreted descriptively and were not considered evidence of causal mediation.

## 2.2. Tasks, models, and interventions

Modular arithmetic was studied over $\mathbb{Z}_p$ for primes $p \in \{59,97,113\}$ using exact classification accuracy. Each input comprised the three-token sequence (a,b,=) with a dedicated separator token. The target was f(a,b) mod p for addition, subtraction, or multiplication. Learned positional embeddings preserved operand order. Standard experiments used a 30/70 train/test split over all $p^2$ ordered pairs with partition seed 42 and a separate ten-seed model pool. The base transformer was set with model dimension 128, four attention heads, MLP dimension 512, and learned positional embeddings. Full-batch AdamW was used with a learning rate of $6 \times 10^{-4}$. The baseline omitted LayerNorm, RMSNorm, residual projection, and used randomly initialized learned token embeddings, linear unembedding at logit scale 1, and weight decay 1.0. The complete Sphere+Fourier configuration used selected sine/cosine token features, unit-norm projection after embedding and each residual operation, cosine unembedding at logit scale 10, and zero weight decay.

The one-layer factorial excluded Fourier initialization and independently varied hard residual projection (N), readout weight normalization (C), logit scale 10 versus 1 (T), and weight decay 1 versus 0 (W). C=1 normalized each output-weight row without separately normalizing the final hidden vector. The readout represented cosine similarity only when N=1 also produced a unit-norm final hidden vector. Hidden magnitude remained in the logits when N=0. Logit scale multiplied the readout as an inverse-temperature. All 16 combinations were evaluated using ten model seeds. Matched contrasts held the C, T, and W operations fixed. Since projection could alter hidden geometry and output scale, ratios conditional on successful runs were not interpreted as marginal causal effects. The fixed-horizon validation and matched-age representation experiments were conducted on NVIDIA GeForce RTX 4090 GPUs. Deterministic algorithms were enabled. The manifests contained hashes for the training code, configurations, results, histories, and checkpoints. Software versions and runtime settings are provided in **Section S1.4** (**Supplementary Material**).

## 3. Results

### 3.1. Component and factorial decomposition

The factorial analysis resolved the interactions underlying the effect of projection (**Fig. 1**; **Section S2.2, Supplementary Material**). Projected models did not produce crossings at either logit scale in the absence of both readout weight normalization and decay (N1C0T0W0 and N1C0T1W0). Grokking was also observed without projection. With weight decay, the unprojected configurations N0C0T1W1 and N0C1T0W1 reached the criterion in 10/10 runs at 21,900 ± 7,331 and 24,680 ± 9,036 epochs, respectively. Combining readout weight normalization, logit scale 10, and decay without projection reduced the mean to 9,420 ± 1,656 epochs with 10/10 crossings. Hidden magnitude was retained in these unprojected readouts.

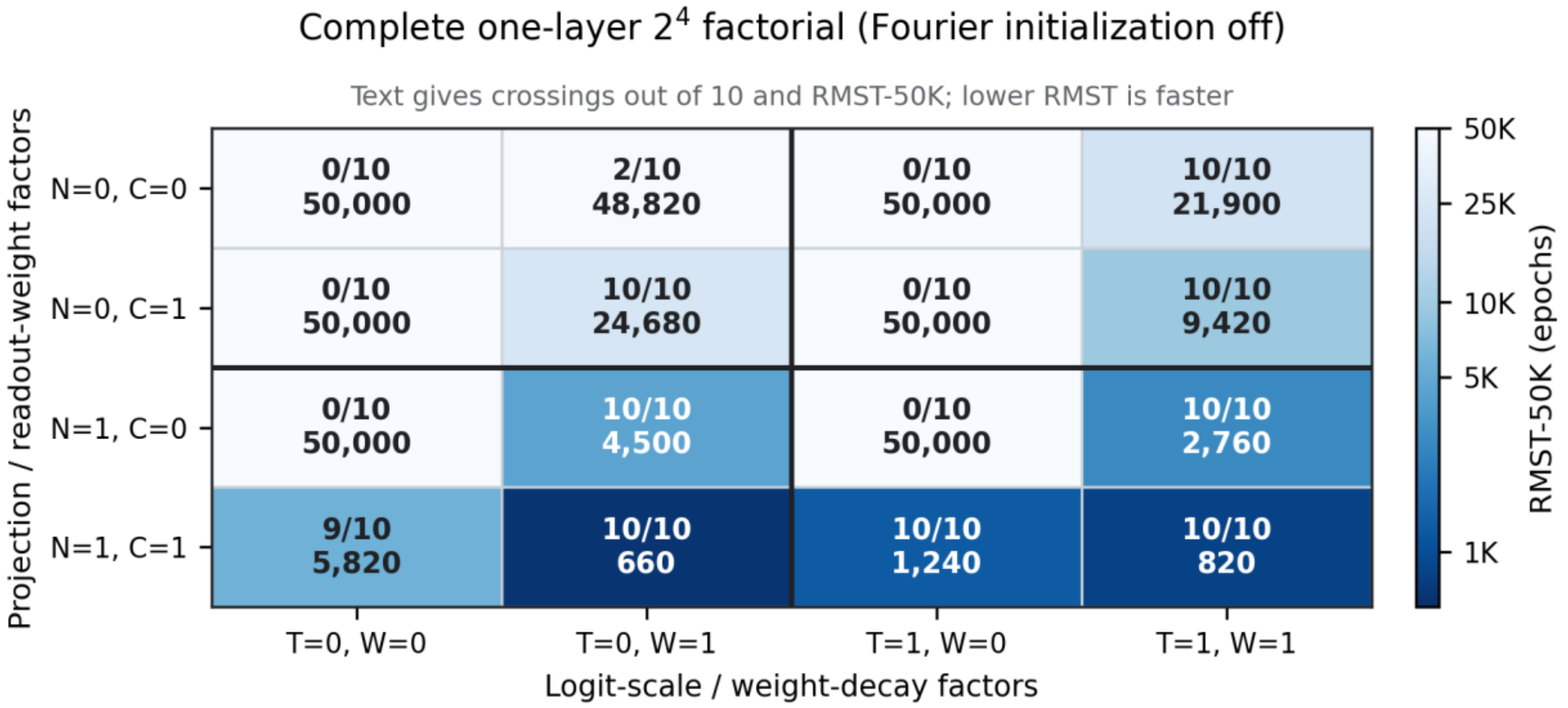

**Fig. 1.** Complete one-layer $2^4$ factorial analysis of $\mathbb{Z}_{113}$ addition with Fourier initialization disabled. Cell color represents RMST-50K. Annotations indicate crossings out of ten and RMST in epochs. Lower RMST values indicate faster crossing. The lowest RMST values occurred with projection. However, no projected runs crossed by 50K when both readout weight normalization and decay were absent.

Nevertheless, hard projection substantially accelerated configurations capable of grokking. Across the four W=1 settings with a minimum of one crossing in both projection states, the successful-run mean without projection was 8-37 times the corresponding mean with projection (**Table 1**). The first contrast compared two successful unprojected runs with ten projected runs. Each of the remaining three contrasts compared ten runs with ten runs. Projection was present in all six configurations with conditional means below 5,000 epochs. Accordingly, the influence of the unit-norm constraint on grokking depended on the accompanying readout and decay settings.

**Table 1:** Matched factorial contrasts with a crossing in both projection states. W=1 throughout and Fourier initialization was disabled. C denotes normalization of readout weights and T denotes logit scale 10 rather than 1. Ratios were calculated by dividing the no-projection mean by the projection mean. Final hidden-vector normalization remained part of N.

| C | T | No projection: crossings, epoch | Projection: crossings, epoch | Conditional ratio |
|---|---|---|---|---|
| 0 | 0 | 2/10, 44,100 ± 6,081 | 10/10, 4,500 ± 1,025 | 9.8× |
| 0 | 1 | 10/10, 21,900 ± 7,331 | 10/10, 2,760 ± 295 | 7.9× |
| 1 | 0 | 10/10, 24,680 ± 9,036 | 10/10, 660 ± 190 | 37.4× |
| 1 | 1 | 10/10, 9,420 ± 1,656 | 10/10, 820 ± 175 | 11.5× |

Component ablation demonstrated that Fourier initialization was insufficient to explain the acceleration (**Section S2.1, Supplementary Material**). The baseline reached grokking criterion in 2/10 runs at 44,100 ± 6,081 epochs among successful runs. Fourier initialization without decay did not produce any crossings by 50K. With decay, 8/10 runs reached the criterion at 28,550 ± 7,906 epochs. Projection with cosine readout without Fourier initialization produced 10/10 crossings at 1,240 ± 363 epochs. Adding Fourier initialization reduced the mean to 720 ± 215 epochs with 10/10 crossings. Thus, Fourier initialization improved this configuration without being required for rapid grokking. The corresponding no-Fourier condition in the factorial was N1C1T1W0.

## 3.2. Independent-partition validation of the complete effect

The 100-run independent-partition study evaluated the robustness of the complete configuration, rather than repeating the individual-factor contrasts (Fig. 2). At one layer, Sphere+Fourier reached the criterion in all 25 runs compared with 8/25 baseline runs (**Table 2**). The paired restricted-time advantage was 45,184 ± 7,327 epochs with a two-factor bootstrap 95% interval of [39,480, 49,128]. At two layers, both configurations reached the criterion in 25/25 runs. Sphere+Fourier generalized earlier with a paired advantage of 16,760 ± 5,307 epochs [13,552, 19,936]. The partition-specific mean advantage was positive at both depths for every partition. These results demonstrated consistent shallow-model acceleration across the five predeclared data partitions.

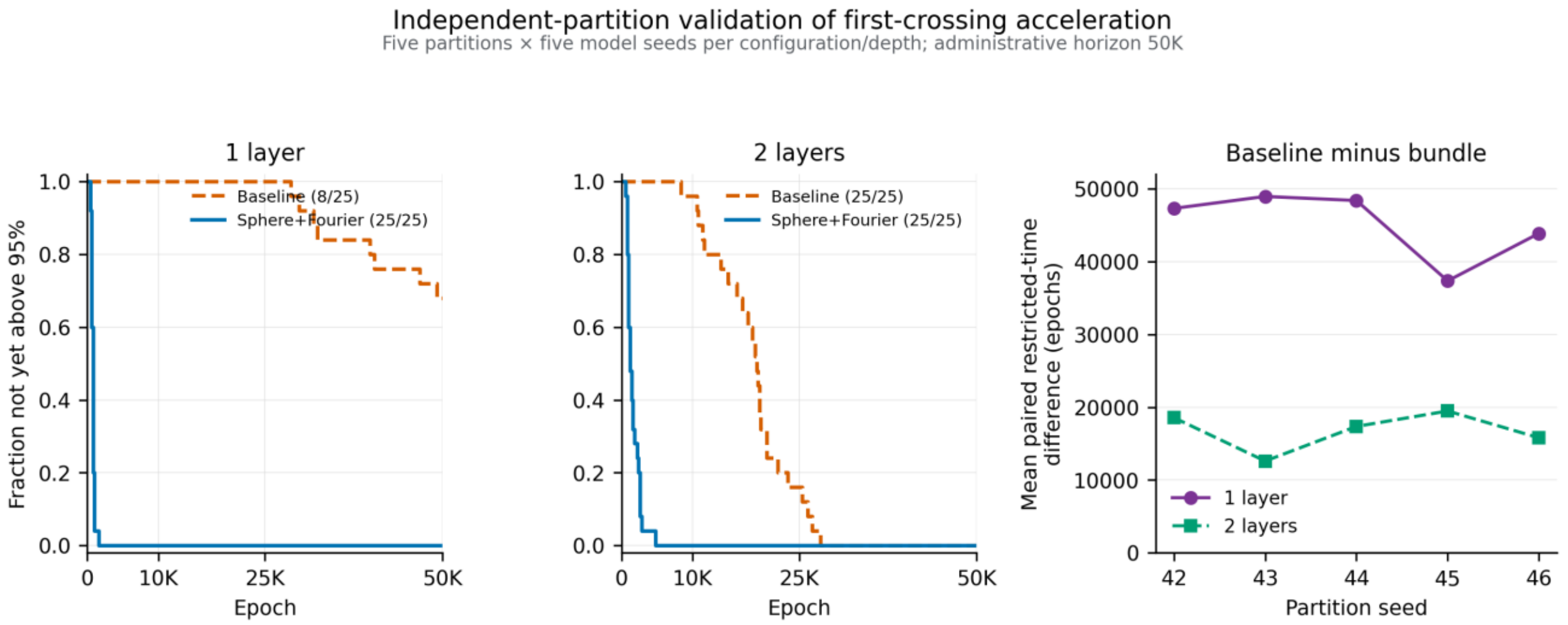

**Fig. 2.** Fixed-horizon validation of the complete baseline and Sphere+Fourier configurations. The first two panels show the fraction of runs not yet observed above 95% held-out accuracy. The third panel shows the matched baseline-minus-Sphere+Fourier RMST-50K difference within each partition. The mean difference was positive for every partition at both depths.

Complete training through to 50K for each run in the study enabled separate assessment of early generalization and subsequent stability. All 50 two-layer runs initially crossed the threshold. Subsequent relapses were recorded for 22/25 baseline runs and 7/25 Sphere+Fourier runs. At epoch 50K, 23/25 baseline runs and 20/25 Sphere+Fourier runs remained above the threshold. Sphere+Fourier generalized earlier and exhibited fewer recorded relapses. However, fewer runs finished above the threshold. This trajectory analysis was specified after observation of the endpoint pattern and remained post hoc. Matched endpoint counts and complete trajectories are provided in **Section S3** (**Supplementary Material**).

**Table 2:** Independent-partition outcomes. Each row includes 25 fixed horizon runs. Conditional epochs were calculated over successful runs. RMST-50K and final accuracy included all runs.

| Configuration and depth | Crossings | Conditional epoch | RMST-50K | Final test accuracy | Above 95% at 50K |
|---|---|---|---|---|---|
| Baseline, 1L | 8/25 | 37,350 ± 7,873 | 45,952 | 44.6 ± 45.3% | 8/25 |
| Sphere+Fourier, 1L | 25/25 | 768 ± 243 | 768 | 100.0 ± 0.01% | 25/25 |
| Baseline, 2L | 25/25 | 18,368 ± 5,303 | 18,368 | 95.0 ± 11.0% | 23/25 |
| Sphere+Fourier, 2L | 25/25 | 1,608 ± 950 | 1,608 | 95.3 ± 11.3% | 20/25 |

## 3.3. Controls, representation evidence, and boundaries

The normalization variants exhibited different behavior under the tested conditions (**Section S2.4**, **Supplementary Material**). With Fourier initialization, normalized readout weights, logit scale 10, zero decay, and the same learning rate, sphere projection produced 10/10 crossings at 720 ± 215 epochs. LayerNorm, RMSNorm, and no normalization produced 0/10 crossings. LayerNorm and RMSNorm contain learned affine scales without enforcing a unit L2 radius. Consequently, these comparisons also altered the magnitude entering the readout and could not isolate normalization from output scale or establish a general advantage of spherical geometry (Singh et al., 2026; Yıldırım, 2026). A separate projected factorial setting produced 10/10 crossings at logit scales 1, 2, 5, and 10 with mean crossing times of 580-820 epochs. Five random five-frequency sets on $\mathbb{Z}_{113}$ each produced 10/10 crossings with means of 760-880 epochs. The results indicated a modest benefit of Fourier initialization without identifying the tested frequency set as a unique explanation for acceleration.

A separate 20-run study compared token embeddings at matched training ages (**Section S4, Supplementary Material**). At 50K, baseline embeddings reached an effective dimensionality of 22.9 ± 6.1 compared with 80.2 ± 0.4 for Sphere+Fourier. The matched-seed difference was 57.3 ± 5.9. This statistic quantified the distribution of variance across embedding directions (Section S1.4, Supporting Information). Matching training ages avoided comparison of an early-stopped fast model with a substantially older baseline. However, differences between the complete configurations and their outcome distributions prevented isolation of hard projection or attribution of grokking to effective dimensionality. Rapid grokking occurred without collapse of this token-embedding statistic. Substantial baseline compression was also insufficient for reliable grokking in these experiments.

Task and depth sweeps identified conditions under which the configuration became unreliable (**Section S5, Supplementary Material**). Sphere+Fourier produced 10/10 crossings for $\mathbb{Z}_{113}$ multiplication and $\mathbb{Z}_{97}$ addition compared with 1/10 for $\mathbb{Z}_{113}$ subtraction and 0/10 for $\mathbb{Z}_{59}$ addition. The operation comparisons lacked matched baselines. The same reference Fourier frequencies were used across primes. Uniform attention permitted grokking on the three-token addition task and accelerated the baseline. Reliability also depended on network depth (**Fig. 3**). In the separate depth sweep, Sphere+Fourier generalized rapidly in 10/10 one-layer runs and 9/10 two-layer runs. Success rates varied at greater depths with 0/10 crossings by 50K at 10, 11, 12, and 16 layers. Baseline configurations continued to cross. These failures could not be attributed specifically to projection since the sweep compared complete configurations.

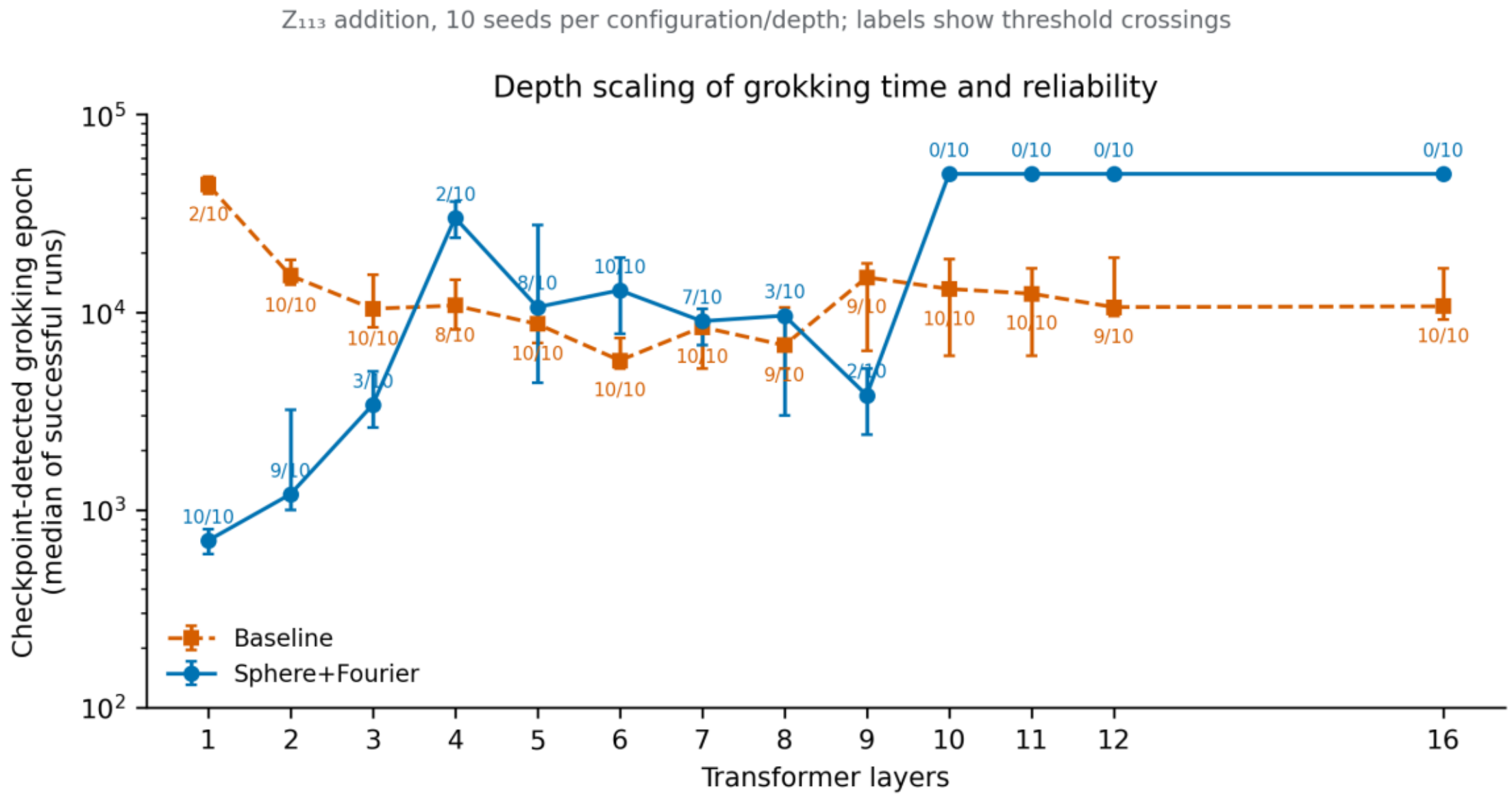

**Fig. 3.** Depth scaling of the complete baseline and Sphere+Fourier configurations on $\mathbb{Z}_{113}$ addition. Points show successful-run medians with IQR. Labels indicate crossings out of ten. Sphere+Fourier configurations with no observed crossings are shown at 50K.

## 4. Discussion

The factorial analysis demonstrated that the influence of projection depended on the readout and training settings. Projection was present in every rapidly generalizing configuration. However, no projected runs reached the criterion in the absence of both readout weight normalization and decay. Weight decay also supported slower grokking without projection. Accordingly, the unit-norm constraint should be evaluated in conjunction with the readout and regularization settings.

The interventions acted on distinct model components. Residual projection constrained hidden vectors, weight decay acted on parameters, and readout weight normalization fixed classifier-row lengths. Projection additionally normalized the final hidden vector. The projection contrasts therefore could not separate geometry from the magnitude entering the readout. The factorial analysis identified interactions among these operations without identifying the learned circuit or explaining why an individual run generalized. Related work distinguished weight magnitude from direction (Truong, 2026a) and demonstrated the influence of feature content and intervention timing on representational priors (Howe, 2026). Preliminary evidence of smooth drift in a squared-loss setting also indicated that a single transition mechanism should not be assumed (Basak, 2026).

The independent-partition study demonstrated shallow-model acceleration across five training subsets. Continued training to a fixed horizon distinguished first crossing, absence of subsequent relapses, and final performance above the threshold. Early stopping would have obscured these differences. The two-layer Sphere+Fourier configuration improved the first two outcomes without improving the third. Previous studies demonstrated continued representation evolution after grokking, changes in optimizer rankings under sustained criteria, and loss of generalization in some Muon-trained transformers after grokking (Truong, 2026b; Wang, 2026; Janati et al., 2026). Accordingly, future confirmatory studies should preregister both first-crossing and sustained or final-accuracy outcomes.

Interpretation of the supporting controls was limited by differences in the interventions and measurements. The tested soft penalty differed from successful soft-constraint methods in target radius, intervention site, readout, and optimization (Section S2.3, Supporting Information). This comparison could not establish superiority of hard constraints (Tiwari et al., 2026). The token-embedding statistic also differed from the hidden-state effective dimensionality controlled by Kazanskii (2026). Neither comparison established a mechanism for the factorial result.

The experiments were limited to small modular-arithmetic transformers. Most factorial configurations used one learning rate, prime, operation, and depth. The partition study evaluated five splits and five model seeds but without the adjustment of its bootstrap for every exploratory comparison. The 200-epoch evaluation grid may have missed brief accuracy changes, and the persistence analysis remained post hoc. A fixed RTX 4090 environment did not establish cross-hardware equivalence (Ootani, 2026). Targeted extensions should vary learning rate in selected factorial settings, isolate projection across depth, match hard and soft constraints by radius and readout, and vary Fourier frequencies and their number at $p=59$. Transfer to language or vision requires separate evaluation.

## 5. Conclusion

Over the prescribed 960 controlled runs, projection accelerated grokking in shallow modular transformers through interactions with readout and regularization settings. It was neither sufficient nor required for generalization within the tested conditions. The complete Sphere+Fourier configuration generalized earlier across five train/test partitions, although fewer of its two-layer runs than baseline runs finished above the accuracy threshold. Its reliability also varied across tasks and depths. Together, these experiments identify component combinations that support rapid grokking and characterize the robustness and limits of the complete Sphere+Fourier configuration.

## Data and code availability

Code, experimental configurations, run-level results, training histories, and analysis scripts are stored in the project repository. The repository is currently private with access provided on demand.

## Disclosure of AI assistance

The authors used OpenAI's ChatGPT and Codex (including GPT-5.5, GPT-5.6, and GPT-6 Astra) and Anthropic's Claude Opus 4.6 to assist with code development, analysis, and manuscript drafting and refinement. The authors reviewed and verified the final methods, results, and conclusions and take full responsibility for the work.

# Supplementary Material:
# Dissecting Geometric Constraints in Grokking: Interaction-Dependent Acceleration in Modular Arithmetic

**Zachary A. Hampton[1], Wasiq Khan[2] (SMIEEE, Ph. D), and Bilal Khan[1], Ph.D.**

[1]School of Computer Science and Engineering, California State University, San Bernardino, {Zachary.Hampton, Bilal.Khan}@csusb.edu

[2]School of Computer Science and Mathematics, Liverpool John Moores University, UK, w.khan@ljmu.ac.uk

## S1. Experimental and reporting details

### S1.1. Tasks, data, architecture, and interventions

Modular arithmetic was studied over $\mathbb{Z}_p$ for primes p ∈ {59, 97, 113}. Each input was represented as (a, b, =) with separator token id p and target *f(a,b) mod p* for f ∈ {+, −, ×}. Learned positional embeddings distinguished (a,b) from (b,a), including for subtraction. Held-out performance was evaluated using exact multiclass accuracy, defined as the fraction of test examples for which the argmax residue equaled the target. All experiments used 30% of the $p^2$ ordered pairs for training and 70% for testing. Partition seed 42 fixed dataset membership in the standard experiments. The fixed-horizon validation crossed partition seeds {42,43,44,45,46}, model seeds {42,137,256,314,555}, depths {1,2}, and two configurations to produce 100 runs without early stopping. Model and optimizer randomness were controlled through separate model seeds. The standard ten-seed pool was {42, 137, 256, 314, 555, 777, 1024, 1337, 2718, 9001}. The base transformer had model dimension 128, four attention heads, MLP dimension 512, learned token and positional embeddings, and one layer unless otherwise specified. The depth experiments evaluated 1-12 and 16 layers. Optimization was performed using full-batch AdamW with a learning rate of $6 \times 10^{-4}$, $\beta = (0.9, 0.999)$, and $\varepsilon = 10^{-8}$. Each epoch comprised one update over the complete training set. The standard baseline omitted LayerNorm, RMSNorm, residual projection, and used randomly initialized learned token embeddings, linear unembedding at logit scale 1, and weight decay 1.0. The standard Sphere+Fourier configuration used Fourier-initialized token embeddings, hard residual projection, cosine unembedding at logit scale 10, and zero weight decay. Comparisons between the complete configurations could not isolate individual components since several properties changed simultaneously.

Hard projection mapped the hidden vector to $h/\|h\|_2$ after embedding and after each attention and MLP residual update. The factorial factor N controlled all projection sites. Factor C normalized each row of the unembedding weight matrix to unit L2 norm without separately normalizing the final hidden vector. The resulting dot product represented cosine similarity only when N=1 also produced a unit-length final hidden vector. Logits retained hidden-vector magnitude when N=0 and C=1. The logits were multiplied by the positive scale τ as an inverse-temperature, not divided by τ. In the binary factorial, T=1 denoted logit scale 10 and T=0 denoted logit scale 1. W=1 denoted weight decay 1.0 and W=0 denoted zero decay. Selected token-embedding coordinates received sine/cosine features at reference frequencies {14, 35, 41, 42, 52}. Separate $\mathbb{Z}_{113}$ experiments compared this set with no Fourier features and five fixed random five-frequency sets. An auxiliary attention experiment replaced learned attention with fixed uniform weights on the three-token input.

## S1.2. Experimental groups

The study comprised 960 runs within 13 experimental groups (**Table S1**). Each run had a corresponding result record and manifest. Conditions repeated in separate run pools were analyzed within their designated experimental groups. Fourier-timing and endpoint-representation experiments were included in the inventory with complete records saved in the repository. The representation analyses reported here used fixed-horizon experiments to compare models at matched training ages.

**Table S1:** Experimental groups and their roles in the study.

| Experimental group | Runs | Design | Primary purpose |
|---|---|---|---|
| Component ablation | 30 | 3 conditions × 10 seeds | Fourier and decay contributions |
| $2^4$ component factorial | 160 | 16 cells × 10 seeds | N/C/T/W interactions |
| Soft-constraint diagnostic | 80 | 8 penalties × 10 seeds | Narrow unmatched diagnostic |
| Operation/prime transfer | 50 | 5 settings × 10 seeds | Boundary conditions |
| Attention routing | 30 | 3 settings × 10 seeds | Learned versus uniform routing |
| Fourier timing | 20 | 2 settings × 10 seeds | Supporting timing evidence |
| Endpoint representation | 20 | 2 settings × 10 seeds | Endpoint-only descriptive comparison |
| Fixed-horizon representation | 20 | 2 configurations × 10 seeds | Matched-age trajectories |
| Depth scaling | 260 | 13 depths × 2 configurations × 10 seeds | Configuration-level depth boundary |
| Normalization variants | 40 | 4 variants × 10 seeds | Sphere/LN/RMS/none control |
| Logit-scale sweep | 80 | 8 logit scales × 10 seeds | Logit-scale robustness |
| Fourier-frequency controls | 70 | 7 initializations × 10 seeds | Frequency-set robustness on $\mathbb{Z}_{113}$ |
| Independent partitions | 100 | 5 partitions × 5 model seeds × 2 depths × 2 configurations | Primary validation |
| Total | 960 | | |

## S1.3. Outcome definitions and statistical reporting

Test accuracy was evaluated at epoch 1 and every 200 epochs thereafter, except in the soft-penalty sweep where evaluation began at epoch 200. Fixed-horizon experiments permanently included epoch 50,000. The grok epoch was defined as the first observed checkpoint with held-out accuracy strictly above 0.95. Crossing times and relapses were resolved only on this evaluation grid. The designation “>50K” indicated that no observed checkpoint crossed the threshold by the administrative horizon. Most standard sweeps employed early stopping only after both training and test accuracy exceeded 0.99 at an evaluation checkpoint. Independent-partition and fixed-horizon representation runs were trained to 50K with early stopping disabled. Conditional grokking time was reported as mean ± sample standard deviation (ddof=1) over successful runs together with the event count. RMST-50K was calculated as the mean of min(T,50,000), assigning 50,000 to runs without an observed crossing. This measure included all runs under the common administrative horizon. Best and final accuracies also included all runs in the specified configuration. Standard deviations were not interpreted as evidence of statistical significance.

The independent-partition study used the paired estimand $\min(T_{baseline},50K) - \min(T_{bundle},50K)$ for matched partition/model seeds. Positive values indicated an advantage for Sphere+Fourier. The two-factor percentile bootstrap independently resampled the five partition seeds and five model seeds with their crossed design preserved. Intervals were calculated using 20,000 replicates and fixed analysis seeds. The persistence analysis was specified after inspection of the completed fixed-horizon independent-partition experiments. Immediate persistence required every observed checkpoint after the first crossing to remain above 0.95. A relapse was defined as any subsequent observed checkpoint at or below 0.95. Observed sustained crossing was defined as the earliest above-threshold checkpoint after which all remaining observed checkpoints through 50K stayed above the threshold. Horizon persistence required final test accuracy above 0.95. A relapsing run was classified as test-only when no post-crossing training-accuracy checkpoint was at or below 0.95. These definitions applied to the recorded evaluation grid without replacing the preregistered first-crossing outcome.

### S1.4. Fixed-horizon representation and runtime protocol.

The canonical one-layer baseline and Sphere+Fourier configurations were each trained for 50,000 epochs using the standard ten seeds with early stopping disabled. Post-update checkpoints were saved at epochs 1, 200, 600, 1,000, 5,000, 10,000, 25,000, 40,000, and 50,000. Token-embedding measurements excluded the separator row. Effective dimensionality was defined as $\exp(H(q))$, where q represented the normalized vector of centered squared singular values. Variance ranks were the minimum centered-SVD ranks required to explain 90% or 99% of variance. Attention entropy was calculated as natural-log entropy averaged over held-out examples, heads, and query positions. Fourier fraction of variance explained (FVE) was calculated using either the five highest-energy frequencies or the reference set. Norm CV denoted the coefficient of variation of token-embedding row norms and did not describe projected residual-stream norms.

The fixed-horizon and independent-partition experiments used Linux x86_64, CPython 3.12.13, NumPy 2.2.6, PyTorch 2.7.1+cu128, CUDA 12.8, cuDNN 9.7.1, and an NVIDIA GeForce RTX 4090. Deterministic algorithms were enabled. TF32 and cuDNN benchmarking were disabled. The manifests contained hashes for the training code, configurations, results, histories, and checkpoints.

## S2. Complete component and control evidence

### S2.1. Component ablation

Fourier initialization was insufficient for rapid grokking under the tested conditions (Table S2). Further attribution required the complete factorial analysis since these conditions differed in multiple components.

**Table S2:** Component ablation for one-layer $\mathbb{Z}_{113}$ models. Conditional epochs were calculated over successful runs. RMST-50K includes all ten runs.

| Condition | Crossings | Conditional epoch | RMST-50K | Best test accuracy |
|---|---|---|---|---|
| Baseline + decay | 2/10 | 44,100 ± 6,081 | 48,820 | 37.5 ± 42.4% |
| Fourier, no decay | 0/10 | >50K | 50,000 | 0.9 ± 0.1% |

| Condition | Crossings | Conditional epoch | RMST-50K | Best test accuracy |
|---|---|---|---|---|
| Fourier + decay | 8/10 | 28,550 ± 7,906 | 32,840 | 89.0 ± 26.3% |
| Sphere, no Fourier | 10/10 | 1,240 ± 363 | 1,240 | 99.3 ± 0.4% |
| Sphere + Fourier | 10/10 | 720 ± 215 | 720 | 99.7 ± 0.3% |

## S2.2. Full $2^4$ factorial

Table S3 reports all 16 factorial configurations, including those without crossings by 50K. Fourier initialization was disabled throughout this configuration.

**Table S3:** One-layer $\mathbb{Z}_{113}$ factorial analysis with Fourier initialization disabled. N, C, T, and W denote hard projection, readout weight normalization, logit scale 10 rather than 1, and weight decay 1, respectively. Conditional epoch is reported as mean ± sample standard deviation over successful runs. Factor C does not normalize the final hidden vector.

| N | C | T | W | Crossings | Conditional epoch | RMST-50K |
|---|---|---|---|---|---|---|
| 0 | 0 | 0 | 0 | 0/10 | >50K | 50,000 |
| 0 | 0 | 0 | 1 | 2/10 | 44,100 ± 6,081 | 48,820 |
| 0 | 0 | 1 | 0 | 0/10 | >50K | 50,000 |
| 0 | 0 | 1 | 1 | 10/10 | 21,900 ± 7,331 | 21,900 |
| 0 | 1 | 0 | 0 | 0/10 | >50K | 50,000 |
| 0 | 1 | 0 | 1 | 10/10 | 24,680 ± 9,036 | 24,680 |
| 0 | 1 | 1 | 0 | 0/10 | >50K | 50,000 |
| 0 | 1 | 1 | 1 | 10/10 | 9,420 ± 1,656 | 9,420 |
| 1 | 0 | 0 | 0 | 0/10 | >50K | 50,000 |
| 1 | 0 | 0 | 1 | 10/10 | 4,500 ± 1,025 | 4,500 |
| 1 | 0 | 1 | 0 | 0/10 | >50K | 50,000 |
| 1 | 0 | 1 | 1 | 10/10 | 2,760 ± 295 | 2,760 |
| 1 | 1 | 0 | 0 | 9/10 | 911 ± 333 | 5,820 |
| 1 | 1 | 0 | 1 | 10/10 | 660 ± 190 | 660 |
| 1 | 1 | 1 | 0 | 10/10 | 1,240 ± 363 | 1,240 |
| 1 | 1 | 1 | 1 | 10/10 | 820 ± 175 | 820 |

## S2.3. Unmatched soft-constraint diagnostic

All soft penalty runs remained ≤ 0.95 for $\lambda \in \{0.001, 0.01, 0.1, 1, 5, 10, 50, 100\}$ with mean best accuracy ranging from 1.5% to 43.0% and reaching 43.0 ± 5.5% at $\lambda$=100, as illustrated in **Fig. S1**. The hard reference produced 10/10 crossings at 720 ± 215 epochs, demonstrating that the tested unit-radius penalties did not reproduce the behavior of the complete hard configuration. Differences in output geometry prevented interpretation as a one-variable hard-versus-soft comparison. The result neither established a phase transition nor contradicted positive findings obtained with target radius $\sqrt{d}$ and different intervention locations (Tiwari et al., 2026).

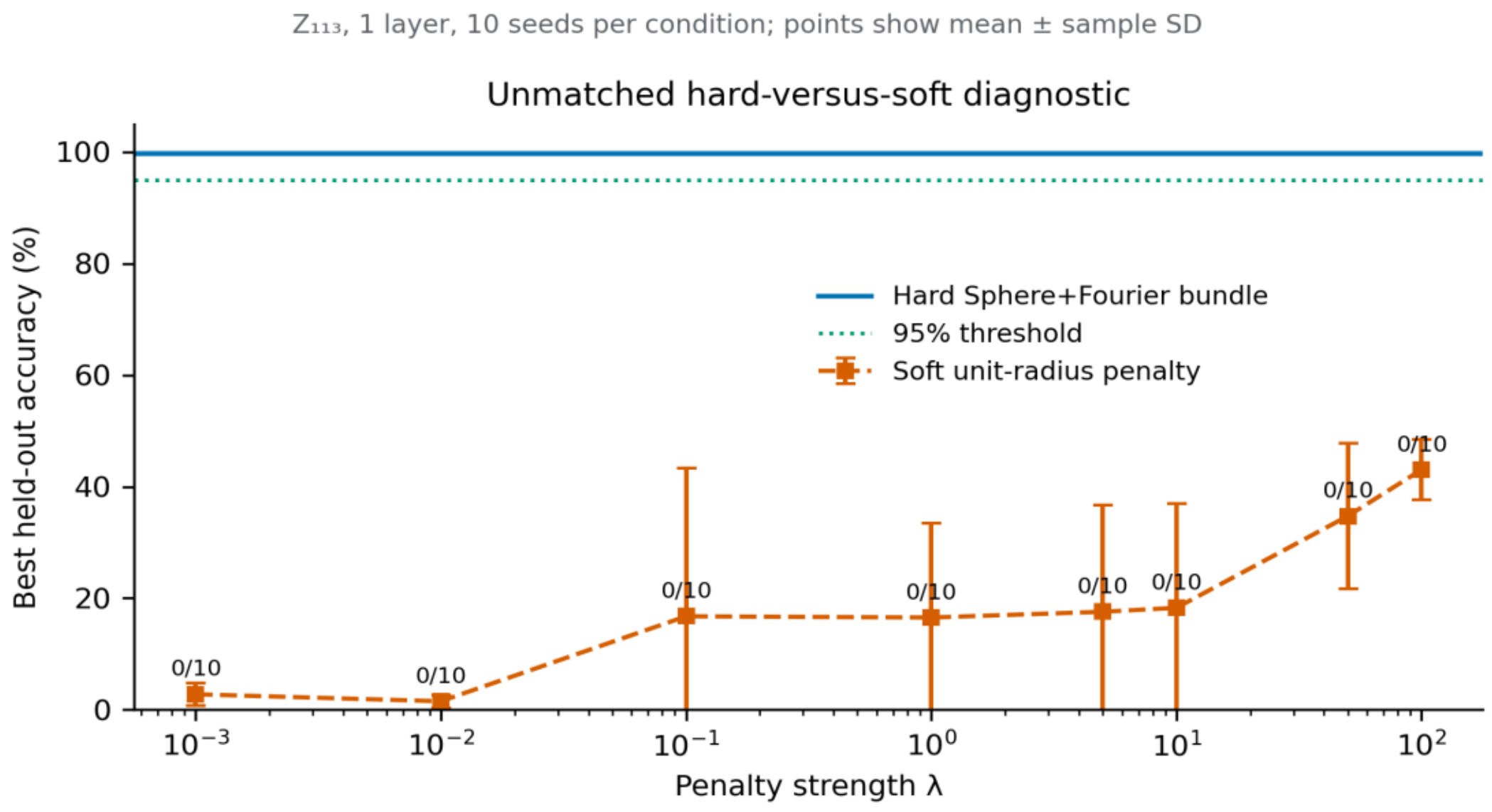

**Fig. S1.** Best held-out accuracy for penalties $\lambda(\|h\|_2-1)^2$. Points show mean ± sample standard deviation across ten seeds. Labels indicate crossings. The hard reference was the complete Sphere+Fourier configuration with different output geometry.

## S2.4. Normalization, logit scale, and frequency controls

**Table S4:** Normalization variants with the same Fourier initialization, readout weight normalization, nominal logit scale, and decay. Hidden radius was not matched.

| Normalization | Crossings | Conditional epoch | RMST-50K | Best test accuracy |
|---|---|---|---|---|
| Sphere | 10/10 | 720 ± 215 | 720 | 99.7 ± 0.3% |
| LayerNorm | 0/10 | >50K | 50,000 | 13.0 ± 6.4% |
| RMSNorm | 0/10 | >50K | 50,000 | 13.9 ± 6.9% |
| None | 0/10 | >50K | 50,000 | 2.3 ± 1.9% |

LayerNorm and RMSNorm normalized the variance or root-mean-square magnitude without enforcing a unit L2 radius. In addition to the learned gains included by both approaches, LayerNorm also included learned offsets which affected output scale and intermediate activations since the readout only normalized its own readout weights. Therefore, the comparison (**Table S4**) could not separate spherical geometry from hidden radius or establish a general normalization advantage. It also did not exclude grokking with LayerNorm or RMSNorm under other placements or hyperparameters (Singh et al., 2026; Yıldırım, 2026).

**Table S5:** Logit-scale sweep with N=1, C=1, W=1, and Fourier initialization disabled. The scale τ multiplies the cosine readout.

| Logit scale τ | Crossings | Conditional epoch | RMST-50K |
|---|---|---|---|
| 0.5 | 10/10 | 1,280 ± 590 | 1,280 |
| 1 | 10/10 | 660 ± 190 | 660 |
| 2 | 10/10 | 580 ± 114 | 580 |
| 5 | 10/10 | 640 ± 126 | 640 |
| 10 | 10/10 | 820 ± 175 | 820 |

| Logit scale τ | Crossings | Conditional epoch | RMST-50K |
|---|---|---|---|
| 20 | 10/10 | 2,000 ± 442 | 2,000 |
| 50 | 10/10 | 5,480 ± 1,051 | 5,480 |
| 100 | 9/10 | 7,422 ± 1,210 | 11,680 |

Mean crossing times were similar across logit scales 1-10 within the resolution of the 200-epoch evaluation grid (**Table S5**). Logit scale 10 was not required for rapid grokking. Larger scales delayed crossing in this sweep.

**Table S6:** Fourier-frequency controls for the one-layer Sphere+Fourier configuration on $\mathbb{Z}_{113}$.

| Initialization | Crossings | Conditional epoch | RMST-50K |
|---|---|---|---|
| No Fourier features | 10/10 | 1,240 ± 363 | 1,240 |
| Reference frequency set | 10/10 | 720 ± 215 | 720 |
| Random set 0 | 10/10 | 760 ± 246 | 760 |
| Random set 1 | 10/10 | 820 ± 346 | 820 |
| Random set 2 | 10/10 | 880 ± 235 | 880 |
| Random set 3 | 10/10 | 860 ± 212 | 860 |
| Random set 4 | 10/10 | 780 ± 175 | 780 |

The behavior produced by all random frequency sets was similar to the reference set on $\mathbb{Z}_{113}$ (**Table S6**).

## S3. Complete-configuration validation and stability

Independent-partition validation used five independently sampled partitions and five model seeds for each combination of configuration and depth (**Table S7**). At one layer, matched restricted-time advantages through partitions were 47,320, 48,960, 48,400, 37,360, and 43,880 epochs. At two layers, the corresponding advantages were 18,560, 12,600, 17,360, 19,480, and 15,800 epochs.

**Table S7:** Independent-partition outcomes. Each row contains 25 runs. Final accuracy includes all runs. Horizon persistence requires final test accuracy above 0.95.

| Configuration and depth | First crossings | Conditional first epoch | RMST-50K | Final test accuracy | Persistent at 50K |
|---|---|---|---|---|---|
| Baseline, 1L | 8/25 | 37,350 ± 7,873 | 45,952 | 44.6 ± 45.3% | 8/25 |
| Sphere+Fourier, 1L | 25/25 | 768 ± 243 | 768 | 100.0 ± 0.01% | 25/25 |
| Baseline, 2L | 25/25 | 18,368 ± 5,303 | 18,368 | 95.0 ± 11.0% | 23/25 |
| Sphere+Fourier, 2L | 25/25 | 1,608 ± 950 | 1,608 | 95.3 ± 11.3% | 20/25 |

**Table S8:** Post hoc persistence counts. Immediate persistence requires every subsequent observed checkpoint to remain above threshold. Ever relapsed denotes any later checkpoint at or below threshold.

| Configuration and depth | First crossed | Immediate persistence | Ever relapsed | Persistent at 50K |
|---|---|---|---|---|
| Baseline, 1L | 8/25 | 8/25 | 0/25 | 8/25 |
| Sphere+Fourier, 1L | 25/25 | 24/25 | 1/25 | 25/25 |
| Baseline, 2L | 25/25 | 3/25 | 22/25 | 23/25 |
| Sphere+Fourier, 2L | 25/25 | 18/25 | 7/25 | 20/25 |

Out of the 22 relapsing two-layer baseline runs (**Fig. S2; Table S8**), 19 had at least one post-crossing training-accuracy checkpoint at or below 0.95. Six of the seven relapsing two-layer Sphere+Fourier runs were classified as test-only. Matched endpoints comprised 19 pairs with persistence in both runs, four baseline-only pairs, one Sphere+Fourier-only pair, and one pair without persistence in either run. Five Sphere+Fourier runs finished below the threshold despite final training accuracy of 1.0. The analysis remained descriptive and post hoc.

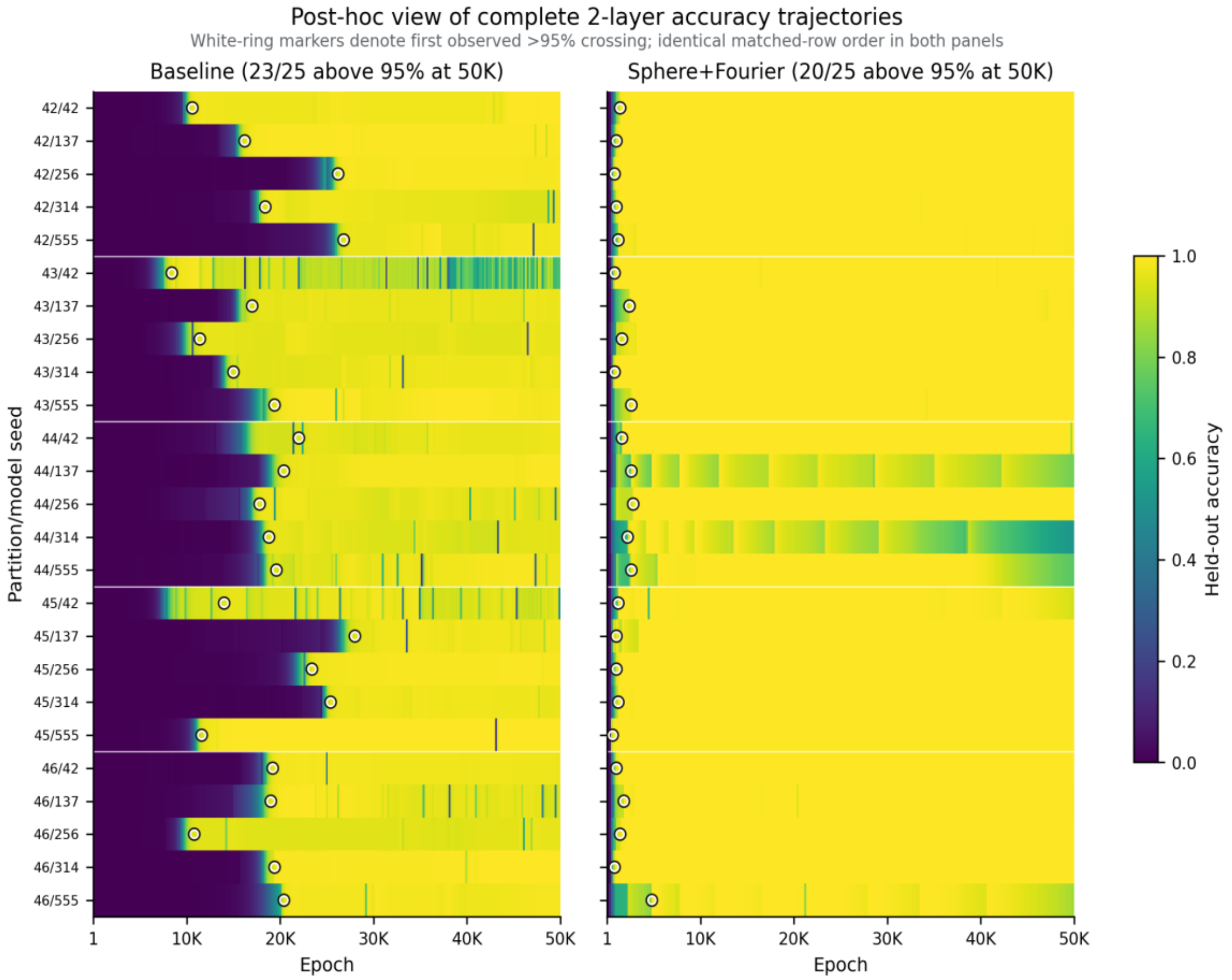

**Fig. S2.** Complete post hoc analysis of two-layer trajectories. Rows follow the matched partition/model-seed order. White-ring markers indicate first observed crossings. Color represents held-out accuracy through 50K. These trajectories provided the basis for the persistence qualification in the main manuscript.

## S4. Matched-age representation evidence

The baseline and Sphere+Fourier configurations began near 72 effective dimensions and remained similar at epoch 1,000 (**Fig. S3**). Endpoint measurements are reported in **Table S9**.

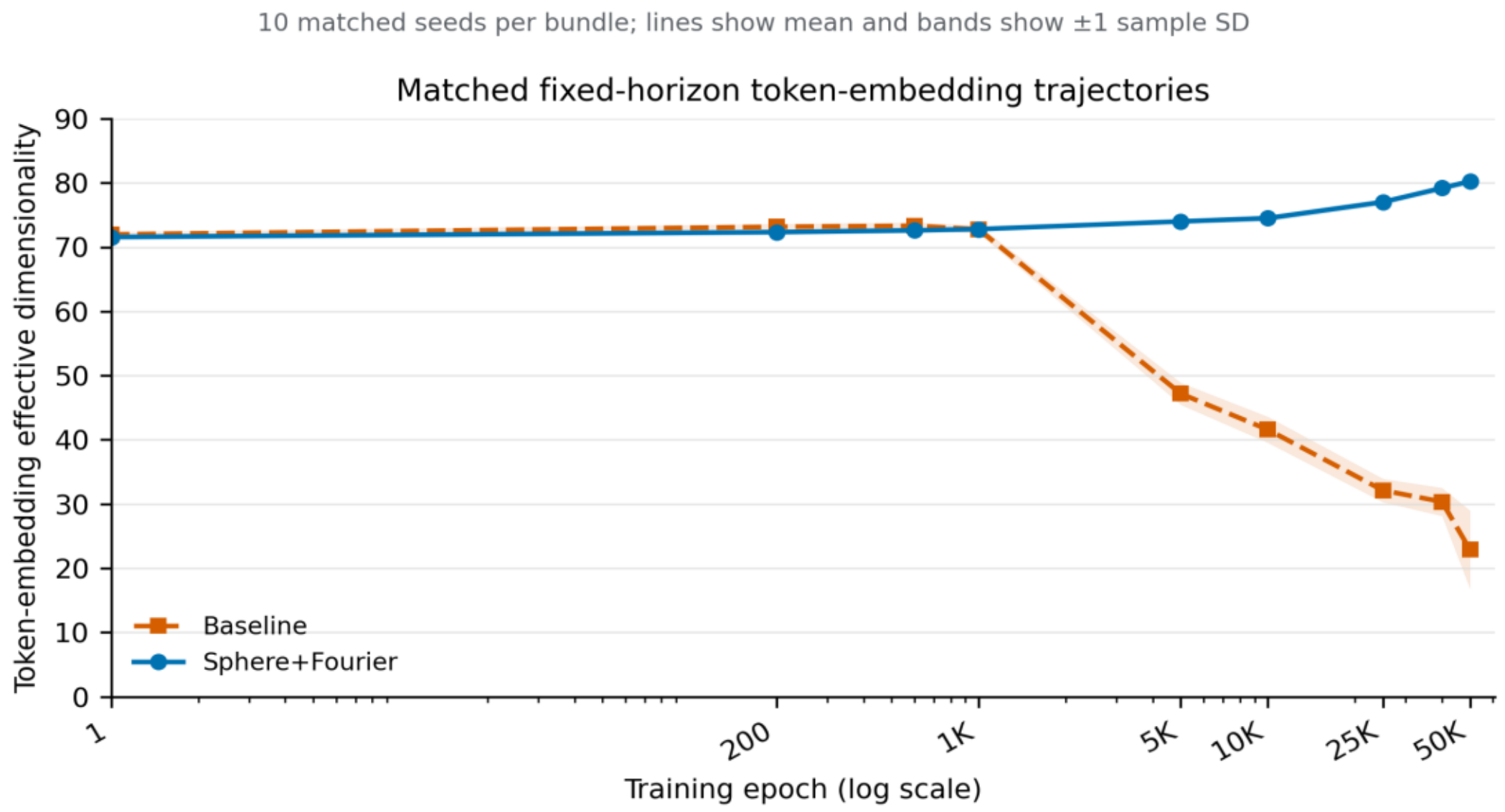

**Fig. S3.** Effective token-embedding dimensionality at matched training ages. Lines indicate means and bands represent ±1 sample standard deviation across ten seeds per configuration. The logarithmic epoch axis does not imply interpolation between saved checkpoints.

The comparison did not isolate projection because configurations differed in initialization, projection, readout, logit scale, and decay. Their outcome distributions were also unequal. Global covariance of the token-embedding table did not represent hidden-state intrinsic dimension, classifier rank, parameter-trajectory dimension, or a causal mediator. The observations remained descriptive.

**Table S9:** Fixed-horizon representation measurements at epoch 50,000. Values are reported as mean ± sample standard deviation over ten seeds. All geometry measurements except attention entropy were calculated from the token-embedding table.

| Metric | Baseline | Sphere+Fourier |
|---|---|---|
| First-crossing success by 50K | 2/10 | 10/10 |
| Test accuracy | 37.5 ± 42.7% | 100.0 ± 0.0% |
| Effective dimensionality | 22.9 ± 6.1 | 80.2 ± 0.4 |
| Rank explaining 90% variance | 19.1 ± 4.8 | 69.1 ± 0.6 |
| Rank explaining 99% variance | 33.3 ± 5.3 | 100.7 ± 0.5 |
| Attention entropy | 0.823 ± 0.071 | 0.975 ± 0.006 |
| Fourier FVE, top five | 48.8 ± 25.1% | 15.8 ± 0.6% |
| Fourier FVE, reference set | 10.1 ± 7.8% | 14.9 ± 0.8% |
| Token-embedding norm CV | 0.053 ± 0.008 | 0.056 ± 0.003 |

## S5. Task, prime, attention, and depth boundaries

The multiplication results demonstrated generalization of the Sphere+Fourier configuration to another operation (**Table S10**). A matched multiplication baseline was not included. Subtraction remained unreliable despite preservation of operand order through positional embeddings. The p=59 configuration failed under the tested frequency set. The p=97 configuration generalized successfully. Accordingly, the p=59 result identified a limitation of the tested configuration without establishing intrinsic incompatibility with that prime.

**Table S10:** Task, prime-size, and attention-routing experiments. Each row represents an independent ten-seed pool. Conditional epochs were calculated over successful runs.

| Condition | Crossings | Conditional epoch | RMST-50K | Best test accuracy |
|---|---|---|---|---|
| Sphere+Fourier, $\mathbb{Z}_{113}$ addition | 10/10 | 720 ± 215 | 720 | 99.7 ± 0.3% |
| Sphere+Fourier, $\mathbb{Z}_{113}$ multiplication | 10/10 | 1,080 ± 454 | 1,080 | 99.6 ± 0.3% |
| Sphere+Fourier, $\mathbb{Z}_{113}$ subtraction | 1/10 | 7,200 | 45,720 | 66.4 ± 26.2% |
| Sphere+Fourier, $\mathbb{Z}_{97}$ addition | 10/10 | 2,400 ± 1,668 | 2,400 | 99.4 ± 0.3% |
| Sphere+Fourier, $\mathbb{Z}_{59}$ addition | 0/10 | >50K | 50,000 | 30.0 ± 0.9% |
| Baseline + uniform attention + decay, $\mathbb{Z}_{113}$ addition | 10/10 | 4,880 ± 1,424 | 4,880 | 99.2 ± 0.1% |
| Sphere+Fourier + uniform attention, $\mathbb{Z}_{113}$ addition | 10/10 | 680 ± 140 | 680 | 99.8 ± 0.2% |

Complete depth results are reported in **Table S11**. Baseline configurations reached the criterion at every tested depth, although not all runs were successful. Since the comparison involved complete configurations, the observed failures at greater depth could not be attributed specifically to repeated projection.

**Table S11:** Complete depth-scaling experiments. Each row represents an independent ten-seed configuration. Conditional epochs were calculated over successful runs.

| Depth | Baseline crossings | Baseline conditional epoch | Baseline RMST | Sphere+ Fourier crossings | Sphere+Fourier conditional epoch | Sphere+ Fourier RMST |
|---|---|---|---|---|---|---|
| 1 | 2/10 | 44,100 ± 6,081 | 48,820 | 10/10 | 720 ± 215 | 720 |
| 2 | 10/10 | 16,660 ± 5,812 | 16,660 | 9/10 | 2,378 ± 2,075 | 7,140 |
| 3 | 10/10 | 13,060 ± 6,726 | 13,060 | 3/10 | 3,667 ± 1,222 | 36,100 |
| 4 | 8/10 | 10,650 ± 3,365 | 18,520 | 2/10 | 30,000 ± 8,768 | 46,000 |
| 5 | 10/10 | 9,380 ± 2,885 | 9,380 | 8/10 | 14,675 ± 13,572 | 21,740 |
| 6 | 10/10 | 8,240 ± 5,483 | 8,240 | 10/10 | 14,060 ± 7,906 | 14,060 |
| 7 | 10/10 | 9,660 ± 5,493 | 9,660 | 7/10 | 9,143 ± 2,084 | 21,400 |
| 8 | 9/10 | 8,267 ± 3,784 | 12,440 | 3/10 | 7,533 ± 3,931 | 37,260 |
| 9 | 9/10 | 14,378 ± 10,241 | 17,940 | 2/10 | 3,800 ± 1,980 | 40,760 |
| 10 | 10/10 | 16,620 ± 13,135 | 16,620 | 0/10 | >50K | 50,000 |
| 11 | 10/10 | 14,320 ± 9,909 | 14,320 | 0/10 | >50K | 50,000 |

| Depth | Baseline crossings | Baseline conditional epoch | Baseline RMST | Sphere+ Fourier crossings | Sphere+Fourier conditional epoch | Sphere+ Fourier RMST |
|---|---|---|---|---|---|---|
| 12 | 9/10 | 13,689 ± 6,799 | 17,320 | 0/10 | >50K | 50,000 |
| 16 | 10/10 | 13,080 ± 6,050 | 13,080 | 0/10 | >50K | 50,000 |